\documentclass{article} 
\usepackage{iclr2026_conference}
\usepackage{times}
\usepackage{microtype}
\usepackage{hyperref}
\usepackage{url}
\usepackage{booktabs}
\usepackage{graphicx}
\usepackage{xcolor}
\usepackage{multirow}
\usepackage{svg}
\usepackage{tikz}
\usepackage{lineno}
\usepackage{amsmath,amsfonts,amssymb}
\usepackage{nicefrac}
\usepackage{todonotes}
\usepackage{wrapfig}
\usepackage{subcaption}
\usepackage{colortbl}
\usepackage{algorithm}
\usepackage{algpseudocode}
\usepackage{multicol}
\usepackage{array}
\usepackage{pgfplots}
\usepgfplotslibrary{fillbetween}
\usepgfplotslibrary{groupplots}
\usepgfplotslibrary{statistics}
\usepackage{pgfplotstable}
\pgfplotsset{compat=1.18}
\usepackage{amsthm}
\usepackage{tcolorbox}
\newtcolorbox{problem}[1][]{
  colback=gray!5,    
  colframe=black!75, 
  fonttitle=\bfseries,
  title=Problem,
  #1
}

\definecolor{darkblue}{rgb}{0, 0, 0.5}
\hypersetup{colorlinks=true, citecolor=darkblue, linkcolor=darkblue, urlcolor=darkblue}

\definecolor{myblue}{rgb}{.12, 0.53, .90}
\definecolor{darkorange}{rgb}{.9, .5, .1}
\definecolor{lightblue}{rgb}{.8, 1, 1}
\definecolor{lightorange}{rgb}{.9, .5, .1}

\definecolor{darkgreen}{rgb}{0.0, 0.31, 0.37}
\definecolor{purple}{rgb}{0.46, 0.41, 0.68}
\definecolor{lightgreen}{rgb}{0.22, 0.78, 0.65}
\definecolor{lightblue}{rgb}{0.22, 0.91, .99}

\newcolumntype{g}{>{\columncolor{lightblue}}c}
\newcolumntype{P}[1]{>{\centering\arraybackslash}p{#1}}

\iclrfinalcopy

\title{SEQR: Secure and Efficient QR-based \\ LoRA Routing}

\author{William Fleshman \& Benjamin Van Durme  \\
Johns Hopkins University\\
\texttt{will.fleshman@jhu.edu} \\
}

\newcommand{\arrow}{{\sc Arrow}}
\newcommand{\spectr}{{\sc SpectR}}
\newcommand{\lag}{{\sc Lag}}
\newcommand{\seqr}{{\sc Seqr}}
\newcommand{\x}{\mathbf{x}}

\newcommand{\y}{\mathbf{y}}
\newtheorem{theorem}{Theorem}[section]
\newtheorem{corollary}{Corollary}[theorem]

\begin{document}

\maketitle

\begin{abstract}
Low-Rank Adaptation (LoRA) has become a standard technique for parameter-efficient fine-tuning of large language models, enabling large libraries of LoRAs, each for a specific task or domain. Efficiently selecting the correct LoRA adapter for a given input remains a challenge, particularly in secure environments where supervised training of routers may raise privacy concerns. Motivated by previous approaches, we formalize the goal of unsupervised LoRA routing in terms of activation norm maximization, providing a theoretical framework for analysis. We demonstrate the discriminative power of activation norms and introduce SEQR, an unsupervised LoRA routing algorithm designed to maximize efficiency while providing strict routing guarantees. SEQR provably identifies the norm-maximizing adapter with significantly greater efficiency, making it a highly scalable and effective solution for dynamic LoRA composition. We validate our results through experiments that demonstrate improved multi-task performance and efficiency.
\end{abstract}

\section{Introduction}

Language model users can benefit from fine-tuning existing models on custom data, but may be constrained by security policies surrounding data access control or retention \citep{adapterswap, flexolmo}. Low-Rank Adaptation (LoRA) \citep{lora} is a popular parameter-efficient technique for fine-tuning these models. Widely-used software packages, such as \texttt{peft} \citep{peft}, and model repositories, such as \textit{huggingface} \citep{huggingface}, have contributed to the proliferation of LoRA-based experts fine-tuned for various tasks or data domains \citep{lotsofloras, lorahub}. The broad deployment of language models has led to techniques for securing and controlling training data \citep{adapterswap, unlearning, flexolmo}. For example, {\sc AdapterSwap} leverages LoRA adapters to segment data into separate parameter groups, enabling user-based access control at the model level \citep{adapterswap}. The authorized LoRAs for a particular user can then be applied to the model at inference time, and adapters can be quickly retrained if training data is later removed to meet retention polices \citep{adapterswap}. 

Naively applying all authorized LoRAs to a model can lead to parameter interference, significantly reducing the model performance \citep{modelsoups, chronopoulou-etal-2023-adaptersoup, taskvector, adapterswap}. Numerous model merging strategies have been developed to address this challenge \citep{ortiz, ties, tang, mario, knots}. Alternatively, LoRAs for the same model can be treated as a \textit{mixture-of-experts} \citep{moe, switch} by learning to route inputs to a smaller set of appropriate adapters \citep{pfeiffer-etal-2021-adapterfusion,wang-etal-2022-adamix, caccia2023multihead, ponti-etal-2023-combining,adapterswap, lorahub,zadouri2024pushing}. Multi-LoRA frameworks have also been used for federated learning, where LoRA training dynamics suggest that the LoRA $A$ matrices learn global features which can be shared among the different adapters \citep{ffalora, fedsalora}.

\begin{figure}[t]
\includegraphics[width=\textwidth]{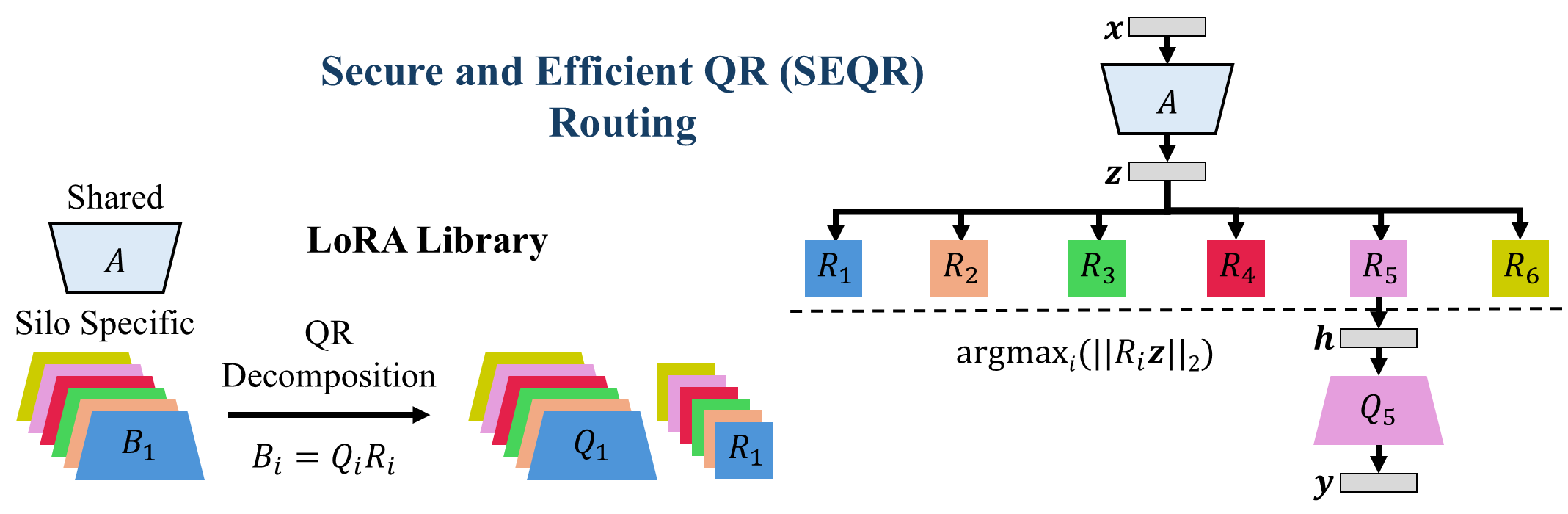}
\caption{Secure and Efficient QR (SEQR) routing: Rank-$r$ LoRAs are trained on multiple datasets using a shared $A$ matrix frozen at initialization. Each $B_i$ is stored in terms of its QR decomposition. During inference, input vectors are routed efficiently using the smaller $r \times r$ matrices.}
\label{fig:seqr}
\end{figure}

Supervised training of a router using data across protected silos is not an option in strict data security scenarios, as adversarial techniques exist for leaking information related to the data \citep{membership, carlini,loraleak, llmleak}. Recent approaches perform LoRA routing in an unsupervised manner by selecting adapters for a given input without any router training or cross-silo data requirements \citep{arrow, lag, spectr}. In this work, we formalize the goal of these techniques and analyze their routing procedures. We introduce a new method, \seqr{} (\autoref{fig:seqr}), which is more efficient than previous approaches while providing strict routing guarantees. Specifically we:
\begin{itemize}
    \item Formalize unsupervised LoRA routing as activation norm-maximization;
    \item Provide theoretical results for current approaches under this framework;
    \item Introduce a more efficient routing scheme, \seqr{}, which provably selects the norm-maximizing adapter; and
    \item Perform empirical experiments demonstrating the benefits of our approach.
\end{itemize}

\section{Background and Related Work}

\subsection{LoRA} LoRA updates the pretrained layer weights $W_0 \in \mathbb{R}^{m \times n}$ by freezing the existing weights and injecting two low-rank matrices of learnable parameters $A \in \mathbb{R}^{r \times n}$ and $B \in \mathbb{R}^{m \times r}$ such that the new weights are $W = W_0 + BA$, with a small rank $r \ll \mathrm{min}(m,n)$ that considerably reduces the number of trainable parameters \citep{lora}. For an input vector $\x \in \mathbb{R}^n$, the output $\y \in \mathbb{R}^m$ can be computed directly with the new weights as $\y = W\x$ or separately as $\y = W_0\x + BA\x$. LoRA routing is necessary in the case where many LoRAs are trained on different groups of data, resulting in a set of available LoRAs $\mathcal{A} = \{B_1A_1, B_2A_2, \dots, B_NA_N\}$ for each adapted layer of the model. The goal of unsupervised routing is to choose the LoRA(s) best suited for each vector in a sequence, without explicitly training the router \citep{arrow, spectr, lag}.

\subsection{Privacy \& Security}

Organizations may have various security or privacy concerns depending on the data used for training individual LoRAs. Training with differential privacy (DP) provides probabilistic guarantees that an adversary can not infer if particular examples were in the training data \citep{diffp, dpsgd}. DP can be used to protect user privacy in cases where adversaries may have access to the LoRA weights \citep{flexolmo}. Stricter security requirements incorporate data access control, completely preventing user access to LoRA weights trained on data the user is unauthorized to view \citep{adapterswap}. In these cases, training a router to distinguish between LoRAs would introduce security concerns, as adversaries with access to the router could potentially leak information from the LoRAs themselves \citep{membership, carlini,loraleak, llmleak}. We focus on the strict security case, where unsupervised routing approaches are needed.

\subsection{Activation Norms}

Unsupervised LoRA routing can be framed as an \textit{in-distribution} (ID) detection problem, where inputs are routed to the adapters trained on data similar to the queries. Prior work has shown that the norm of the activation vector produced by model layers can effectively distinguish between in- and out-of distribution (OOD) data \citep{featurenorm, nac, normamp, massivellm, oodnap}. ID data tends to produce large activation spikes in neural networks, including in large language models \citep{massivellm}. \citet{featurenorm} analyze this phenomenon and find that the activation norm distinguishes OOD and ID similar to a classifier confidence score. These findings justify trying to route to LoRAs which maximize the norm of adapter activations $\lvert\lvert BA\x\rvert\rvert$.

\subsection{Arrow Routing}

\citet{arrow} use the singular value decomposition (SVD) to convert each LoRA adapter $B_iA_i \in \mathcal{A}$ into a product of three matrices with an equivalent product:
\begin{equation}
    B_iA_i = U_iS_iV_i^T, 
    \label{eq:svd}
\end{equation}
where $U_i \in \mathbb{R}^{m \times r}$ is the orthonormal matrix of left singular vectors, $S_i \in \mathbb{R}^{r \times r}$ is the diagonal matrix of singular values, and $V_i\in \mathbb{R}^{n \times r}$ is the orthonormal matrix of right singular vectors. \arrow{} routing leverages the fact that the right singular vector $\mathbf{v}_i$ associated with the largest singular value corresponds to the direction capturing the most variation in the space of input vectors $\x$ \citep{arrow}. This \textit{arrow vector} $\mathbf{v}_i$ satisfies
    $\mathbf{v}_i = \mathrm{max}_{\x,\lvert\lvert\x\rvert\rvert_2 =1}\lvert\lvert B_iA_i\x\rvert\rvert_2$,
meaning it maximizes the norm of the corresponding adapter activations among unit-length input vectors. We use norm-maximization as the explicit goal in this work, allowing for analysis of these approaches. \citet{arrow} use the set of arrows as prototypes for each of the adapters in $\mathcal{A}$, assigning the most weight to the adapter corresponding to the arrow satisfying $\mathrm{argmax_i}\lvert \mathbf{v}_i^T\x\rvert$. The use of vector prototypes makes \arrow{} routing especially efficient, requiring a simple dot product per adapter: $\mathcal{O}(Nn)$ for $N$ adapters with input dimension $n$. \arrow{} routing performs reasonably well, and the authors empirically show that the ID adapter tends to produce higher \arrow{} scores \citep{arrow}.

\subsection{Spectral Routing and LAG} \spectr{} builds on \arrow{} by using all right singular vectors to make routing decisions \citep{spectr}. \autoref{eq:svd} is used by \spectr{} to convert each adapter into two new matrices:
\begin{equation}
    \hat{B}_i = U_i
    \label{spectrB}
\end{equation}
\begin{equation}
    \hat{A}_i = S_iV_i^T,
    \label{spectrA}
\end{equation} such that $\hat{B}_i\hat{A}_i = B_iA_i$ with $\hat{A}_i$ now containing the orthogonal directions of maximum variation scaled by the singular values. \spectr{} generalizes the \arrow{} scoring method by assigning the most weight to the adapter satisfying $\mathrm{argmax}_i\lvert\lvert \hat{A}_i\x\rvert\rvert_2$. Computing the {\sc SpectR} routing scores is less efficient than \arrow{}: $\mathcal{O}(Nrn)$, but \spectr{} outperforms \arrow{} in routing accuracy and downstream task performance \citep{spectr}. 

LoRA-Augmented Generation (\lag) combines the efficiency of \arrow{} routing with the improved performance of \spectr{} by using a two-stage approach \citep{lag}. First, \lag{} performs top-$k$ filtering using \arrow{} to reduce the final routing decision to $k\ll N$ adapters. \lag{} then uses \spectr{} to route to the top adapter in the filtered set. Routing complexity is reduced to $\mathcal{O}(Nn + krn)$ while still outperforming \arrow{} \citep{lag}.

\subsection{Shared A}

While \arrow{}, \spectr{}, and \lag{} use traditional LoRA fine-tuning, recent work explores a special case of LoRA where the $A$ matrix is frozen at initialization or shared among several LoRAs in a federated setting, resulting in similar or improved performance with reduced storage costs \citep{lorafa, ffalora, asymmetry, fedsalora}. \citet{asymmetry} provides a theoretical analysis showing that the LoRA updates are dominated by the B matrix during fine-tuning, and that a LoRA with a frozen random $A$ matrix should perform similarly to one that is fully trained. The asymmetry in training dynamics lends itself to using a global $A$ matrix and unique $B$ matrices in multi-LoRA scenarios \citep{ffalora,fedsalora}. We explore this direction in our work, and show that a shared $A$ matrix allows for more efficient unsupervised LoRA routing techniques.  

\section{Theoretical Results and SEQR}
\label{theory}

\paragraph{Problem Statement}

We formalize the goal of unsupervised LoRA routing to provide a framework for theoretical analysis. Given the success of using activations for ID/OOD detection and the similar motivation of current unsupervised routing approaches, we propose the following problem: 

\begin{problem}
\textbf{LoRA Activation Norm-Maximization.} Given a library of LoRA adapters, \\$\mathcal{A} = \{B_1A_1, B_2A_2, \dots, B_NA_N\}$ and an input vector $\x$, efficiently find  $\mathrm{argmax}_i\lvert\lvert B_iA_i\x\rvert\rvert_2$.
\end{problem}
We add ``efficiently'' to the problem statement as an algorithm that simply computes all activation norms directly would be $\mathcal{O}(Nr(m+n))$, far worse than current routing approaches. We will demonstrate the discriminative power of LoRA activation norms in \autoref{norms}. 

\subsection{Arrow is not Norm-Maximizing}
Our first result shows that \arrow{} is not guaranteed to find the norm-maximizing adapter.
\begin{theorem}
There exists a set of LoRA adapters $\{B_1A_1, B_2A_2, \dots, B_NA_N\}$ with corresponding arrow vectors $\{v_1, v_2, \dots, v_N\}$ and $\x\in \mathbb{R}^n$ where
$\mathrm{argmax}_i\lvert\mathbf{v}_i^T\x\rvert \ne \mathrm{argmax}_i\lvert\lvert B_iA_i\x\rvert\rvert_2$.
\end{theorem}
We provide the proof by construction in \autoref{construction} and confirm with experiments in \autoref{accuracy}. The main observation from the proof is that alignment with the top singular vector is not enough to guarantee the adapter will have the largest norm, as misalignment can be overcome with larger singular values. Routing with \lag{} inherits the lack of guarantee from \arrow{}, but the top-$k$ selection improves the chances of including the norm-maximizing adapter in the set used for \spectr{} selection. 

\subsection{SpectR is Norm-Maximizing}
Our next results show that \spectr{} scores are equivalent to the activation norms, and therefore \spectr{} is norm-maximizing. The proof for Theorem 3.2 is provided in \autoref{pr:spectr}.

\begin{theorem}
    \label{th:spectr}
    Let $B\in \mathbb{R}^{m \times r}$ and $A \in \mathbb{R}^{r \times n}$ be LoRA matrices with $\hat{A}$ derived from $BA$ using Equations 1 and 3, then $\forall \x \in \mathbb{R}^n$, $\lvert\lvert \hat{A}\x\rvert\rvert_2 = \lvert\lvert BA\x\rvert\rvert_2$.
\end{theorem}

\begin{corollary}
    Let $\{B_1A_1, B_2A_2, \dots, B_NA_N\}$ be a set of LoRA adapters converted with Equations 1-3 to the set $\{\hat{B}_1\hat{A}_1, \hat{B}_2\hat{A}_2, \dots, \hat{B}_N\hat{A}_N\}$, then $\mathrm{argmax}_i\lvert\lvert \hat{A}_i\x\rvert\rvert_2 = \mathrm{argmax}_i\lvert\lvert B_iA_i\x\rvert\rvert_2$.
\end{corollary}

These results show that \spectr{} provides optimal routing under the stated goal. We are interested in new approaches providing the same guarantee but with improved efficiency. 

\subsection{Secure and Efficient QR (SEQR) Routing}
\label{seqr}

Now we explore the special case of our problem statement where all adapters in $\mathcal{A}$ share the same matrix $A$. This matrix is randomly initialized and kept frozen to ensure the same data security provided by other unsupervised routing approaches. For an input $\x$, we compute $\mathbf{z} = A\x$ as an intermediate step. Routing is then required for the set of $B$ matrices. Directly computing the norm for all would require $\mathcal{O}(Nmr)$, which is already equivalent to \spectr{} for $m=n$. We can improve further by doing a one-time preprocessing step similar to the SVD in \arrow{} and \spectr. We precompute the reduced QR decomposition of each $B_i$:
\begin{equation}
    B_i = Q_iR_i,
\end{equation} where $Q_i \in \mathbb{R}^{m \times r}$ is an orthogonal matrix and $R_i \in \mathbb{R}^{r \times r}$ is upper triangular. Similar to \spectr{}, we can throw away the original $B_i$ and store the adapter in this new form. The vector $\mathbf{z}$ is then routed to the adapter satisfying $\mathrm{argmax}_i\lvert\lvert R_i\mathbf{z}\rvert\rvert_2$.\footnote{We z-score these raw scores based on our findings in \autoref{norms}.} The routing complexity is only $\mathcal{O}(Nr^2)$, which is far better than \spectr{} and is even more efficient than \arrow{} routing in the typical LoRA scenario where $r \ll n$. We present the complete \seqr{} routing process in \autoref{alg:seqr}. Like \spectr{}, we show \seqr{} scores are equivalent to the activation norm for each adapter. Therefore, \seqr{} always selects the norm-maximizing adapter. The proof for Theorem 3.3 is provided in \autoref{pr:seqr}.
 
\begin{theorem}
    Let $B\in \mathbb{R}^{m \times r}$ and $A \in \mathbb{R}^{r \times n}$ be LoRA matrices such that $B=QR$ from Equation 4, then $\forall \x \in \mathbb{R}^n$, $\lvert\lvert RA\x\rvert\rvert_2 = \lvert\lvert BA\x\rvert\rvert_2$.
\end{theorem}
\begin{corollary}
     Let $\{B_1, B_2, \dots, B_N\}$ be a set of LoRA adapters with a shared $A$ matrix and $\{Q_1R_1, Q_2R_2, \dots, Q_NR_N\}$ from Equation 4, then $\mathrm{argmax}_i\lvert\lvert R_iA\x\rvert\rvert_2 = \mathrm{argmax}_i\lvert\lvert B_iA_i\x\rvert\rvert_2$.
\end{corollary}

\subsection{Routing Complexity}

We revisit the routing complexities of \arrow{} routing, \spectr{}, \lag{}, and \seqr{} using dimensions reported in the \lag{} experiments for added context \citep{lag}. \autoref{tab:complexity} includes the FLOPs used for routing by each method in this example, including the naive approach of computing the norm directly for each adapter. \seqr{} is two orders of magnitude more efficient than any other approach. \seqr{} also decreases storage costs by offsetting the storage of each $R_i$ by sharing $A$ across the library. \arrow{} can also take advantage of improved storage when using a shared $A$ matrix, but arrow vectors require more space than the $R_i$ matrices when $n>r^2$.

\begin{algorithm}[t]
\caption{Secure and Efficient QR (SEQR) Routing}
\label{alg:seqr}
\begin{algorithmic}
\Require 
    Pretrained weight matrix $W \in \mathbb{R}^{m \times n}$ \\
    Shared adapter matrix $A \in \mathbb{R}^{r \times n}$ 
    \Comment{Randomly initialized and frozen during training}\\
    LoRA matrices $\{B_i \in \mathbb{R}^{m \times r}\}_{i=1}^N$ \\
    Norm statistics $\{\mu_i, \sigma_i\}_{i=1}^N$
    \Comment{Estimated using training data}

\Statex
\Statex \textbf{Preprocessing}
\For{each adapter $B_i$}
    \State Compute reduced QR decomposition: $B_i = Q_i R_i$
    \Comment{$B_i$ can be discarded}
\EndFor

\Statex
\Statex \textbf{Inference} (given input $\x \in \mathbb{R}^n$)
\State Compute shared intermediate representation: $\mathbf{z} \gets A\x$

\For{each adapter $i = 1, \dots, N$}
    \State Projected activation: $\mathbf{h}_i \gets R_i \mathbf{z}$
    \State Score: $s_i \gets (\lVert \mathbf{h}_i \rVert_2 -\mu_i)/\sigma_i$
    \Comment{Z-scored activation norm}
\EndFor

\State Select top adapter: $i^* \gets \arg\max_i s_i$
\Comment{Adapter with max activation norm}
\State Compute final output: $\mathbf{y} \gets W \x + Q_{i^*} \mathbf{h}_{i^*}$
\Comment{$Q_{i^*}\mathbf{h}_{i^*} = B_{i^*}A\x$}

\Statex
\Return $\mathbf{y}$
\end{algorithmic}
\end{algorithm}

\begin{table}[h]
\caption{Routing complexity and example FLOPs for each method assuming $N=1000$ adapters, $n=m=4096$ hidden dimension, $k=20$ \lag{} filtering, and $r=8$ rank adapters.}
\label{tab:complexity}
\centering
\small
\begin{tabular}{r|c|c|c|c|c}
    & Naive & \spectr{} & \lag{} & \arrow{} & \seqr{} \\
    \toprule
    \textbf{FLOPs} & 66M & 33M & 5M & 4M & \textbf{64}K \\
    \textbf{Complexity} & $\mathcal{O}(Nr(m+n))$ & $\mathcal{O}(Nrn)$ & $\mathcal{O}(Nn + krn)$ & $\mathcal{O}(Nn)$ & $\mathcal{O}(Nr^2)$\\
\end{tabular}
\end{table}

\section{Experiments}
\label{experiments}

We conduct experiments to validate our theoretical results and to test whether \seqr{} provides similar or better performance over less efficient alternatives. First, we confirm prior work showing that using a fixed $A$ matrix in LoRA works as well as learning $A$ individually. We analyze the differences in activation norms between these two settings and introduce a calibration step to ensure norms between adapters are on the same scale. We measure the ability of each approach to select the norm-maximizing adapter and the resulting multi-task performance and efficiency.

\subsection{Models and Data}
\label{models}

We replicate the experiments of \citet{spectr} using the Llama-3.2-3B-Instruct model \citep{llama}. We train LoRAs for a variety of tasks: agnews\footnote{\url{http://groups.di.unipi.it/~gulli/AG_corpus_of_news_articles.html}}, cola \citep{warstadt-etal-2019-neural}, dbpedia \citep{dbpedia}, hellaswag \citep{zellers-etal-2019-hellaswag}, mnli \citep{williams-etal-2018-broad}, mrpc \citep{dolan-brockett-2005-automatically}, qnli \citep{rajpurkar-etal-2016-squad}, qqp,\footnote{\url{https://quoradata.quora.com/First-Quora-Dataset-Release-Question-Pairs}} rte \citep{wang-etal-2018-glue}, and sst2 \citep{socher-etal-2013-recursive}. Similar to \citet{arrow}, we subsample the datasets for computational feasibility. Using different random seeds, we produce three sets of LoRAs per dataset and category (shared vs. unique $A$ matrix), each trained on 1000 samples from the corresponding dataset. Learning rates were optimized per dataset and category but shared across random seeds. All evaluations are performed using a held-out set of 1000 examples from each dataset. LoRA $A$ matrices are initialized from $\mathcal{N}(0, 1/r^2)$ and frozen in the shared setting. The $B$ matrices are initialized with 0s and trained in both cases \citep{lora}. The complete adapter training details are included in \autoref{adapters}. 

\subsection{Unique vs. Shared}

Before measuring routing performance, we ensure that using frozen $A$ matrices results in similar LoRA performance. \autoref{tab:oracle} shows the accuracy of each adapter on the corresponding test set, averaged across the three different initializations. Accuracy is within 1\% between the two categories in most cases, with the largest deviation being a 2\% difference on hellaswag when using the frozen $A$ matrices. Overall, the average performance is nearly identical, a finding consistent with prior work showing similar performance with a frozen $A$ \citep{lorafa, ffalora, asymmetry}.

\begin{table}[t]
    \caption{Accuracy for LoRAs using a unique or fixed $A$ matrix shared across datasets. }
    \label{tab:oracle}
    \centering
    \small
    \begin{tabular}{c|c|c|c|c|c|c|c|c|c|c|c}
        & agnews & cola & dbped & hswag & mnli & mrpc & qnli & qqp & rte & sst2 & AVG \\
        \toprule
        Unique & 90.4 & 78.8 & 98.7 & 83.6 & 86.1 & 84.7 & 84.9 & 86.5 & 88.2 & 92.4 & 87.4\\
        Shared & 90.0 & 78.9 & 99.0 & 81.5 & 85.7 & 85.0 & 85.5 & 86.3 & 87.9 & 92.8 & 87.3\\
    \end{tabular}
\end{table}

\subsection{Activation Norms}
\label{norms}

Activation norms of a given adapter can be informative for distinguishing ID from OOD data. However, to ensure bias-free routing, these norms must be comparable across adapters. For instance, the agnews adapter may produce lower norms than the cola adapter regardless of the dataset, even if it generates higher norms on agnews data specifically. In such cases, the routing procedure would be biased toward selecting the cola adapter. We explore and mitigate this potential bias in norms.

\begin{figure}[!h]
\centering
\begin{tikzpicture}
\begin{groupplot}[
    group style={
        group size=2 by 1,    
        horizontal sep=1.5cm, 
        ylabels at=edge left,
        xlabels at=edge bottom
    },
    width=0.48\columnwidth, 
    height=5cm,
    ylabel={Activation Norm},
    ymin=0.0,
    ymax=1.4,
    xtick={1,2,3,4,5,6,7,8,9,10},
    xticklabels={agnews, cola,dbpedia,hswg,mnli,mrpc,qnli,qqp,rte,sst2},
    xticklabel style={rotate=45, anchor=east, font=\small, xshift=0.4em, yshift=-0.4em},
    grid=major,
    grid style={solid,gray!30},
]
\nextgroupplot[xlabel={Unique $A$ Matrix}]

    \addplot+[boxplot prepared={median=0.14, lower quartile=0.10, upper quartile=0.2, lower whisker=0.01, upper whisker=0.56}, blue, boxplot/draw direction=y, line width=1pt, solid] coordinates {};
    \addplot+[boxplot prepared={median=0.14, lower quartile=0.06, upper quartile=0.24, lower whisker=0.01, upper whisker=0.68}, blue, boxplot/draw direction=y, line width=1pt, solid] coordinates {};
    \addplot+[boxplot prepared={median=0.14, lower quartile=0.09, upper quartile=0.24, lower whisker=0.01, upper whisker=0.65}, blue, boxplot/draw direction=y, line width=1pt, solid] coordinates {};
    \addplot+[boxplot prepared={median=0.14, lower quartile=0.07, upper quartile=0.22, lower whisker=0.01, upper whisker=1.13}, blue, boxplot/draw direction=y, line width=1pt, solid] coordinates {};
    \addplot+[boxplot prepared={median=0.15, lower quartile=0.09, upper quartile=0.25, lower whisker=0.01, upper whisker=0.58}, blue, boxplot/draw direction=y, line width=1pt, solid] coordinates {};
    \addplot+[boxplot prepared={median=0.15, lower quartile=0.08, upper quartile=0.23, lower whisker=0.01, upper whisker=0.62}, blue, boxplot/draw direction=y, line width=1pt, solid] coordinates {};
    \addplot+[boxplot prepared={median=0.16, lower quartile=0.08, upper quartile=0.25, lower whisker=0.01, upper whisker=0.77}, blue, boxplot/draw direction=y, line width=1pt, solid] coordinates {};
    \addplot+[boxplot prepared={median=0.15, lower quartile=0.08, upper quartile=0.24, lower whisker=0.01, upper whisker=0.67}, blue, boxplot/draw direction=y, line width=1pt, solid] coordinates {};
    \addplot+[boxplot prepared={median=0.16, lower quartile=0.09, upper quartile=0.25, lower whisker=0.01, upper whisker=1.06}, blue, boxplot/draw direction=y, line width=1pt, solid] coordinates {};
    \addplot+[boxplot prepared={median=0.17, lower quartile=0.09, upper quartile=0.3,  lower whisker=0.01, upper whisker=0.92}, blue, boxplot/draw direction=y, line width=1pt, solid] coordinates {};

\nextgroupplot[xlabel={Shared $A$ Matrix}]

    \addplot+[boxplot prepared={median=0.03, lower quartile=0.01, upper quartile=0.04, lower whisker=0.0, upper whisker=0.07}, orange, boxplot/draw direction=y, line width=1pt, solid] coordinates {};
    \addplot+[boxplot prepared={median=0.16, lower quartile=0.08, upper quartile=0.21, lower whisker=0.01, upper whisker=0.39}, orange, boxplot/draw direction=y, line width=1pt, solid] coordinates {};
    \addplot+[boxplot prepared={median=0.64, lower quartile=0.29, upper quartile=0.89, lower whisker=0.06, upper whisker=1.28}, orange, boxplot/draw direction=y, line width=1pt, solid] coordinates {};
    \addplot+[boxplot prepared={median=0.15, lower quartile=0.08, upper quartile=0.2,  lower whisker=0.01, upper whisker=0.31}, orange, boxplot/draw direction=y, line width=1pt, solid] coordinates {};
    \addplot+[boxplot prepared={median=0.29, lower quartile=0.15, upper quartile=0.39, lower whisker=0.02, upper whisker=0.64}, orange, boxplot/draw direction=y, line width=1pt, solid] coordinates {};
    \addplot+[boxplot prepared={median=0.29, lower quartile=0.14, upper quartile=0.37, lower whisker=0.02, upper whisker=0.61}, orange, boxplot/draw direction=y, line width=1pt, solid] coordinates {};
    \addplot+[boxplot prepared={median=0.3,  lower quartile=0.14, upper quartile=0.4,  lower whisker=0.02, upper whisker=0.65}, orange, boxplot/draw direction=y, line width=1pt, solid] coordinates {};
    \addplot+[boxplot prepared={median=0.17, lower quartile=0.08, upper quartile=0.22, lower whisker=0.01, upper whisker=0.42}, orange, boxplot/draw direction=y, line width=1pt, solid] coordinates {};
    \addplot+[boxplot prepared={median=0.29, lower quartile=0.11, upper quartile=0.37, lower whisker=0.02, upper whisker=0.61}, orange, boxplot/draw direction=y, line width=1pt, solid] coordinates {};
    \addplot+[boxplot prepared={median=0.31, lower quartile=0.17, upper quartile=0.41, lower whisker=0.02, upper whisker=0.79}, orange, boxplot/draw direction=y, line width=1pt, solid] coordinates {};

\end{groupplot}
\end{tikzpicture}
\caption{Distribution of average activation norms for each dataset when using LoRA adapters with unique $A$ matrices or a fixed $A$ matrix shared across adapters.}
\label{fig:boxplot}
\end{figure}
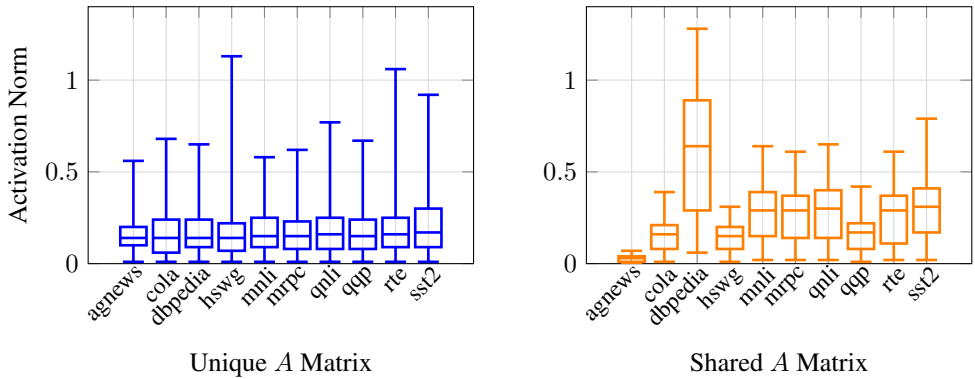

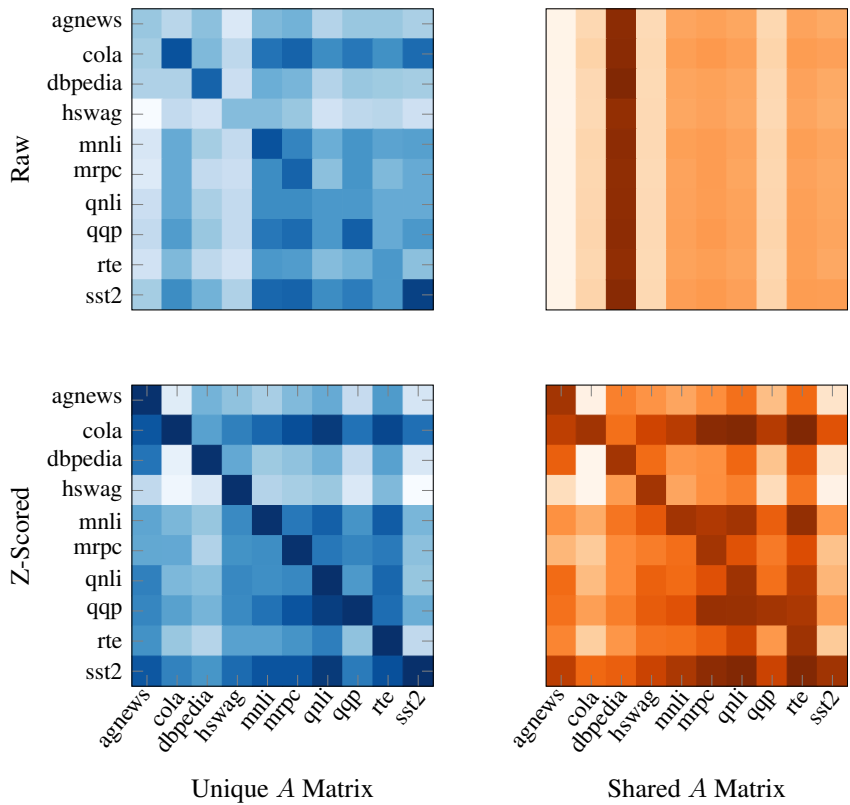
\begin{figure}[t]
    \centering
\pgfplotsset{colormap={Blues}{
rgb255(0cm)=(247,251,255)
rgb255(1cm)=(222,235,247)
rgb255(2cm)=(198,219,239)
rgb255(3cm)=(158,202,225)
rgb255(4cm)=(107,174,214)
rgb255(5cm)=(66,146,198)
rgb255(6cm)=(33,113,181)
rgb255(7cm)=(8,81,156)
rgb255(8cm)=(8,48,107)
},
colormap={Oranges}{
rgb255(0cm)=(255,245,235)
rgb255(1cm)=(254,230,206)
rgb255(2cm)=(253,208,162)
rgb255(3cm)=(253,174,107)
rgb255(4cm)=(253,141,60)
rgb255(5cm)=(241,105,19)
rgb255(6cm)=(217,72,1)
rgb255(7cm)=(166,54,3)
rgb255(8cm)=(127,39,4)
}}

\begin{tikzpicture}
\begin{groupplot}[
  group style={
    group size=2 by 2,
    horizontal sep=1.5cm,
    vertical sep=1cm,
    ylabels at=edge left,
    xlabels at=edge bottom
  },
  width=.4\columnwidth,
  height=.4\columnwidth,
  view={0}{90},
  enlargelimits=false,
  axis on top,
  point meta=explicit,
  xtick={0,...,9},
  xticklabels={agnews,cola,dbpedia,hswag,mnli,mrpc,qnli,qqp,rte,sst2},
  x tick label style={rotate=50,anchor=east, xshift=0.5ex, yshift=-0.5ex},
  ytick={0,...,9},
  y dir=reverse,
  yticklabels={agnews,cola,dbpedia,hswag,mnli,mrpc,qnli,qqp,rte,sst2},
  ticklabel style={font=\small},
]

\nextgroupplot[point meta min=0.138, point meta max=0.20, ylabel={Raw}, xtick={\empty}, colormap name=Blues]
\addplot [
  matrix plot*,
  mesh/cols=10
] table [meta=z] {
x y z
0 0 0.162
1 0 0.156
2 0 0.164
3 0 0.147
4 0 0.166
5 0 0.168
6 0 0.157
7 0 0.162
8 0 0.162
9 0 0.159
0 1 0.160
1 1 0.192
2 1 0.166
3 1 0.155
4 1 0.184
5 1 0.188
6 1 0.178
7 1 0.183
8 1 0.177
9 1 0.186
0 2 0.158
1 2 0.158
2 2 0.188
3 2 0.152
4 2 0.169
5 2 0.167
6 2 0.157
7 2 0.162
8 2 0.161
9 2 0.160
0 3 0.138
1 3 0.154
2 3 0.150
3 3 0.165
4 3 0.165
5 3 0.162
6 3 0.150
7 3 0.155
8 3 0.156
9 3 0.151
0 4 0.148
1 4 0.170
2 4 0.160
3 4 0.154
4 4 0.192
5 4 0.180
6 4 0.169
7 4 0.176
8 4 0.172
9 4 0.173
0 5 0.146
1 5 0.170
2 5 0.154
3 5 0.152
4 5 0.178
5 5 0.188
6 5 0.164
7 5 0.176
8 5 0.166
9 5 0.170
0 6 0.152
1 6 0.170
2 6 0.159
3 6 0.154
4 6 0.178
5 6 0.178
6 6 0.175
7 6 0.175
8 6 0.170
9 6 0.170
0 7 0.154
1 7 0.174
2 7 0.162
3 7 0.154
4 7 0.183
5 7 0.186
6 7 0.175
7 7 0.189
8 7 0.170
9 7 0.175
0 8 0.150
1 8 0.166
2 8 0.155
3 8 0.150
4 8 0.175
5 8 0.174
6 8 0.165
7 8 0.168
8 8 0.175
9 8 0.164
0 9 0.160
1 9 0.178
2 9 0.168
3 9 0.158
4 9 0.187
5 9 0.188
6 9 0.178
7 9 0.182
8 9 0.175
9 9 0.196
};

\nextgroupplot[point meta min=0.02, point meta max=0.65, ytick={\empty}, xtick={\empty}]
\addplot [
  matrix plot*,
  mesh/cols=10
] table [meta=z] {
x y z
0 0 0.027
1 0 0.149
2 0 0.624
3 0 0.141
4 0 0.277
5 0 0.288
6 0 0.274
7 0 0.150
8 0 0.282
9 0 0.267
0 1 0.027
1 1 0.167
2 1 0.629
3 1 0.147
4 1 0.293
5 1 0.306
6 1 0.290
7 1 0.161
8 1 0.298
9 1 0.289
0 2 0.026
1 2 0.148
2 2 0.646
3 2 0.143
4 2 0.278
5 2 0.286
6 2 0.273
7 2 0.149
8 2 0.284
9 2 0.266
0 3 0.024
1 3 0.145
2 3 0.609
3 3 0.146
4 3 0.273
5 3 0.284
6 3 0.268
7 3 0.146
8 3 0.277
9 3 0.259
0 4 0.026
1 4 0.156
2 4 0.623
3 4 0.145
4 4 0.291
5 4 0.299
6 4 0.282
7 4 0.156
8 4 0.293
9 4 0.278
0 5 0.025
1 5 0.152
2 5 0.616
3 5 0.142
4 5 0.282
5 5 0.299
6 5 0.276
7 5 0.153
8 5 0.284
9 5 0.271
0 6 0.026
1 6 0.155
2 6 0.619
3 6 0.144
4 6 0.284
5 6 0.296
6 6 0.283
7 6 0.155
8 6 0.290
9 6 0.274
0 7 0.026
1 7 0.158
2 7 0.623
3 7 0.145
4 7 0.287
5 7 0.303
6 7 0.285
7 7 0.160
8 7 0.292
9 7 0.278
0 8 0.026
1 8 0.153
2 8 0.614
3 8 0.143
4 8 0.283
5 8 0.293
6 8 0.279
7 8 0.153
8 8 0.291
9 8 0.270
0 9 0.027
1 9 0.162
2 9 0.633
3 9 0.147
4 9 0.294
5 9 0.305
6 9 0.289
7 9 0.160
8 9 0.298
9 9 0.295
};

\nextgroupplot[point meta min=-0.42, point meta max=0.00, xlabel={Unique $A$ Matrix}, , ylabel={Z-Scored}, colormap name=Blues]
\addplot [
  matrix plot*,
  mesh/cols=10
] table [meta=z] {
x y z
0 0 -0.003
1 0 -0.370
2 0 -0.219
3 0 -0.248
4 0 -0.276
5 0 -0.233
6 0 -0.203
7 0 -0.315
8 0 -0.174
9 0 -0.348
0 1 -0.061
1 1 -0.000
2 1 -0.186
3 1 -0.128
4 1 -0.089
5 1 -0.048
6 1 -0.017
7 1 -0.107
8 1 -0.037
9 1 -0.102
0 2 -0.110
1 2 -0.386
2 2 -0.003
3 2 -0.199
4 2 -0.262
5 2 -0.248
6 2 -0.216
7 2 -0.314
8 2 -0.186
9 2 -0.354
0 3 -0.308
1 3 -0.404
2 3 -0.354
3 3 -0.002
4 3 -0.291
5 3 -0.274
6 3 -0.259
7 3 -0.359
8 3 -0.231
9 3 -0.420
0 4 -0.193
1 4 -0.226
2 4 -0.255
3 4 -0.144
4 4 -0.003
5 4 -0.116
6 4 -0.077
7 4 -0.162
8 4 -0.071
9 4 -0.224
0 5 -0.198
1 5 -0.199
2 5 -0.288
3 5 -0.159
4 5 -0.151
5 5 -0.003
6 5 -0.115
7 5 -0.137
8 5 -0.119
9 5 -0.244
0 6 -0.129
1 6 -0.228
2 6 -0.241
3 6 -0.142
4 6 -0.153
5 6 -0.142
6 6 0.003
7 6 -0.171
8 6 -0.088
9 6 -0.254
0 7 -0.138
1 7 -0.185
2 7 -0.221
3 7 -0.145
4 7 -0.106
5 7 -0.058
6 7 -0.022
7 7 -0.003
8 7 -0.098
9 7 -0.208
0 8 -0.158
1 8 -0.257
2 8 -0.292
3 8 -0.186
4 8 -0.186
5 8 -0.162
6 8 -0.125
7 8 -0.248
8 8 0.003
9 8 -0.309
0 9 -0.062
1 9 -0.132
2 9 -0.164
3 9 -0.095
4 9 -0.058
5 9 -0.058
6 9 -0.015
7 9 -0.120
8 9 -0.051
9 9 0.002
};

\nextgroupplot[point meta min=-0.33, point meta max=0.04, xlabel={Shared $A$ Matrix}, ytick={\empty}]
\addplot [
  matrix plot*,
  mesh/cols=10
] table [meta=z] {
x y z
0 0 -0.003
1 0 -0.319
2 0 -0.128
3 0 -0.154
4 0 -0.179
5 0 -0.146
6 0 -0.108
7 0 -0.213
8 0 -0.099
9 0 -0.279
0 1 -0.028
1 1 -0.000
2 1 -0.107
3 1 -0.043
4 1 -0.022
5 1 0.027
6 1 0.035
7 1 -0.020
8 1 0.038
9 1 -0.066
0 2 -0.086
1 2 -0.344
2 2 -0.002
3 2 -0.102
4 2 -0.158
5 2 -0.149
6 2 -0.096
7 2 -0.221
8 2 -0.074
9 2 -0.282
0 3 -0.266
1 3 -0.358
2 3 -0.166
3 3 -0.002
4 3 -0.179
5 3 -0.148
6 3 -0.128
7 3 -0.263
8 3 -0.114
9 3 -0.322
0 4 -0.150
1 4 -0.188
2 4 -0.114
3 4 -0.075
4 4 -0.001
5 4 -0.015
6 4 -0.001
7 4 -0.085
8 4 0.017
9 4 -0.154
0 5 -0.203
1 5 -0.231
2 5 -0.144
3 5 -0.124
4 5 -0.105
5 5 -0.002
6 5 -0.066
7 5 -0.121
8 5 -0.058
9 5 -0.218
0 6 -0.101
1 6 -0.210
2 6 -0.144
3 6 -0.087
4 6 -0.101
5 6 -0.064
6 6 0.003
7 6 -0.108
8 6 -0.022
9 6 -0.202
0 7 -0.109
1 7 -0.170
2 7 -0.125
3 7 -0.080
4 7 -0.065
5 7 0.011
6 7 0.009
7 7 -0.003
8 7 -0.011
9 7 -0.165
0 8 -0.134
1 8 -0.236
2 8 -0.157
3 8 -0.112
4 8 -0.107
5 8 -0.083
6 8 -0.041
7 8 -0.160
8 8 0.003
9 8 -0.230
0 9 -0.026
1 9 -0.097
2 9 -0.085
3 9 -0.039
4 9 -0.011
5 9 0.024
6 9 0.037
7 9 -0.040
8 9 0.037
9 9 0.001
};

\end{groupplot}
\end{tikzpicture}
    \caption{Raw (top) versus z-scored (bottom) activation norms for the adapters using unique (left) or shared (right) $A$ matrices. Rows are datasets and columns are the applied adapter.
    }
    \label{fig:norm}
\end{figure}

We gather the average activation norms across model layers for each adapter in \autoref{fig:boxplot}. We find that the activation norms are very consistent across adapters when using LoRAs trained with unique $A$ matrices. However, when the adapters share a frozen $A$ matrix across datasets, the variance in activation norms is considerable. To address the issue, we introduce an offline calibration step for the norm-based approaches. We compute the mean, $\mu_i$, and standard deviation, $\sigma_i$, of the activation norm for each adapter using its associated training data. The scores for \spectr{} become $s_i = (\lvert\lvert\hat{A}_i\mathbf{x}\rvert\rvert_2 - \mu_i) / \sigma_i$ and similarly for \seqr{} $s_i = (\lvert\lvert R_iA\mathbf{x}\rvert\rvert_2 - \mu_i) / \sigma_i$, which are the z-scores of the original raw scores to ensure all adapters are on the same scale. These normalized scores are already included in \autoref{alg:seqr}. \arrow{} scores remain the same, as $\mathbf{v}_i$ is unit-length by construction.

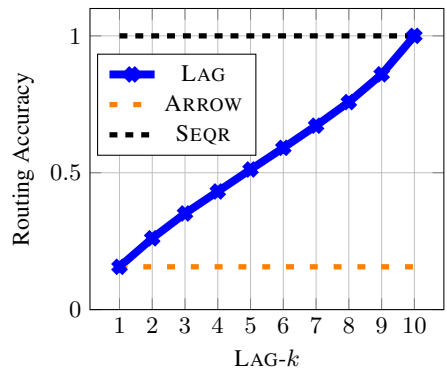
\begin{wrapfigure}[17]{r}{0.47\columnwidth}
    \centering
\centering
\small
\begin{tikzpicture}
\begin{axis}[
    xlabel={\lag{}-$k$},
    ylabel={Routing Accuracy},
    xtick={1,2,3,4,5,6,7,8,9,10},
    ymin=0, ymax=1.1,
    width=.45\columnwidth,
    height=.4\columnwidth,
    legend style={at={(.02,.85)}, anchor=north west},
    grid=both,
    grid style={line width=.1pt, draw=gray!20},
    major grid style={line width=.2pt,draw=gray!50},
]

\addplot[
    color=blue,
    mark=x,
    line width=3pt,
    mark size=3pt,
] coordinates {
    (1,0.15644)
    (2,0.260543)
    (3,0.35131)
    (4,0.431547)
    (5,0.511863)
    (6,0.59091)
    (7,0.671963)
    (8,0.758153)
    (9,0.859587)
    (10,1.0)
};
\addlegendentry{\lag{}}

\addplot[
    color=orange,
    loosely dashed,
    line width=2pt,
] coordinates {(1,0.15644) (10,0.15644)};
\addlegendentry{\arrow{}}

\addplot[
    color=black,
    dashed,
    line width=2pt,
] coordinates {(1,1.0) (10,1.0)};
\addlegendentry{\seqr}

\end{axis}
\end{tikzpicture}
    \caption{Routing accuracy as $k$ increases for \lag{}. \lag{} is equivalent to \arrow{} at $k=1$ and to \spectr{} at $k=10$. \arrow{} chooses the norm-maximizing adapter for 16\% of tokens.}
    \label{fig:accuracy}
\end{wrapfigure}

We visualize the impact of z-scoring by measuring the average activation norm for each adapter, on each dataset, before and after normalizing the scores (\autoref{fig:norm}). We see that the norms for adapters with unique $A$ matrices are already discriminative, but normalizing does sharpen the distribution. The biased norms in the shared case completely prevent accurate discrimination, with the dbpedia adapter producing the largest average norm regardless of the dataset. Z-scoring significantly improves the results, leading to similar relative averages when compared to the traditionally trained LoRAs with unique $A$ matrices. 

\subsection{Routing Accuracy}
\label{accuracy}

We validate our theoretical results by measuring the percentage of tokens routed to the norm-maximizing adapter (\autoref{fig:accuracy}). \arrow{} chooses the adapter with the top singular vector most aligned to the input. This adapter is the norm-maximizing adapter for only 16\% of tokens. The routing accuracy of \lag{} scales almost linearly with $k$, as \lag{} is equivalent to \arrow{} at $k=1$ and equivalent to \spectr{} at $k=10$. \spectr{} and \seqr{} both choose the norm-maximizing adapter in all cases. These results empirically confirm our theoretical findings and are consistent with prior work showing \arrow{} routing accuracies just above random change and improved routing with \spectr{} \citep{spectr}.

\subsection{Task Performance}

We measure multi-task performance by evaluating the routing methods on the withheld data from each dataset. Keeping with previous work, we include {\sc Mu}-routing as an additional baseline \citep{arrow, spectr}. {\sc Mu} forgoes routing to individual LoRAs and instead computes the mean update using all adapters: $\mathbf{y} = W\mathbf{x} + \frac{1}{N}\sum_{i=1}^NB_iA\mathbf{x}$. While simple, averaging adapters can lead to poor performance due to interference in parameter space, especially with a large number of adapters \citep{ortiz,tang, knots}.

\citet{lag} use $k=20$ with \lag, filtering their adapter library to 2\% of the total before using \spectr{} to make the final selection. With only 10 adapters, we use a 30\% reduction with $k=3$ for demonstration purposes, but note the \lag{} task performance is equivalent to \arrow{} for $k=1$ and to \spectr{} and \seqr{} at $k=10$. We control for variation in task difficulty by dividing each score by the performance of the correct adapter from \autoref{tab:oracle} \citep{arrow, spectr}. We report the mean performance and standard deviation over three random seeds in \autoref{tab:performance}. \seqr{} and \spectr{} route equivalently, so we only include \seqr{} in the table. All other approaches significantly outperform {\sc Mu} routing. \seqr{} achieves the highest average score in all cases, outperforming \arrow{} and \lag{}.\footnote{A paired t-test produces $p=0.013$ when comparing with \arrow{} and $p=0.096$ with \lag{}.} The similar task-performance with \lag{} and identical performance with \spectr{} make differences in efficiency a primary consideration for choosing among the various approaches. Next, we explore these differences in more detail.

\begin{table}[t]
    \caption{Mean and standard deviation of performance achieved across datasets and routing methods. \spectr{} achieves identical performance as \seqr{} but at a higher computational cost.}
    \label{tab:performance}
    \centering
    \begin{tabular}{l|r|r|r|r}
    & \multicolumn{1}{c|}{\sc Mu} & \multicolumn{1}{c|}{\arrow} & \multicolumn{1}{c|}{\lag} & \multicolumn{1}{c}{\seqr} \\
    \toprule
    agnews & 16 $\pm$ 9.7 & 89 $\pm$ 0.9 & 89 $\pm$ 1.0 & \textbf{91} $\pm$ 0.5\\
    cola & 86 $\pm$ 3.6 & 92 $\pm$ 1.6 & 94 $\pm$ 0.8 & \textbf{96} $\pm$ 0.9\\
    dbpedia & 89 $\pm$ 2.6 & \textbf{100} $\pm$ 0.1 & \textbf{100} $\pm$ 0.2 & \textbf{100} $\pm$ 0.2\\
    hswag & 53 $\pm$ 0.0 & 78$\pm$12.2 & 86 $\pm$ 2.1 & \textbf{87} $\pm$ 2.3\\
    mnli & 48 $\pm$ 2.5 & 78 $\pm$ 2.1 & \textbf{81} $\pm$ 6.2 & \textbf{81} $\pm$ 7.5\\
    mrpc & 76 $\pm$ 4.5 & 92 $\pm$ 1.2 & \textbf{93} $\pm$ 1.5 & \textbf{93} $\pm$ 2.2\\
    qnli & 79 $\pm$ 0.6 & 92 $\pm$ 2.8 & 95 $\pm$ 0.9 & \textbf{96} $\pm$ 1.4\\
    qqp & 61 $\pm$ 4.4 & \textbf{97} $\pm$ 0.9 & \textbf{97} $\pm$ 0.2 & \textbf{97} $\pm$ 0.1\\
    rte & 73 $\pm$ 1.1 & 94 $\pm$ 3.1 & \textbf{96} $\pm$ 1.5 & \textbf{96} $\pm$ 2.1\\
    sst2 & 94 $\pm$ 0.3 & 97 $\pm$ 1.0 & \textbf{98} $\pm$ 0.9 & \textbf{98} $\pm$ 2.3\\
    \midrule
    AVG & \multicolumn{1}{l|}{67.5} & \multicolumn{1}{l|}{\hspace{4pt}90.9} & \multicolumn{1}{l|}{\hspace{4pt}92.9} & \multicolumn{1}{l}{\hspace{4pt}\textbf{93.5}}
    \end{tabular}
\end{table}

\subsection{Routing Efficiency}

\begin{figure}[t]
    \small
    \centering
    \pgfplotsset{
  log x ticks with fixed point/.style={
      xticklabel={
        \pgfkeys{/pgf/fpu=true}
        \pgfmathparse{exp(\tick)}%
        \pgfmathprintnumber[fixed relative, precision=3]{\pgfmathresult}
        \pgfkeys{/pgf/fpu=false}
      }
  }}

\begin{tikzpicture}
\begin{groupplot}[
  group style={
    group size=2 by 3,
    horizontal sep=2cm,
    vertical sep=1.5cm
  },
  width=6cm, height=4.9cm,
  xlabel style={font=\small},
  ylabel style={font=\small},
  legend style={font=\small, anchor={north west}, at={(.05,.95)}},
  grid=both,
  tick label style={font=\small},
  xminorgrids=false, yminorgrids=false,
  minor tick style={draw=none},
  every axis plot/.append style={
    line width=2pt,
    mark size=2pt,
    mark options={line width=2pt}
  }
]

\nextgroupplot[xlabel=Hidden Dimension, ylabel=FLOPs,ymode=log]
\addplot[color=orange, mark=*] table[
x=hidden_dim, y=flops, col sep=comma,
] {data/arrow_h.csv};
\addplot[color=teal, mark=triangle*] table[
x=hidden_dim, y=flops, col sep=comma,
] {data/lag_h.csv};
\addplot[color=blue, mark=square*] table[
x=hidden_dim, y=flops, col sep=comma,
] {data/seqr_h.csv};
\addplot[color=black, mark=+] table[
x=hidden_dim, y=flops, col sep=comma,
] {data/spectr_h.csv};

\nextgroupplot[xlabel=Hidden Dimension, ylabel=Memory,ymode=log]
\addplot[color=orange, mark=*] table[
x=hidden_dim, y=memory, col sep=comma,
] {data/arrow_h.csv};
\addplot[color=teal, mark=triangle*] table[
x=hidden_dim, y=memory, col sep=comma,
] {data/lag_h.csv};
\addplot[color=blue, mark=square*] table[
x=hidden_dim, y=memory, col sep=comma,
] {data/seqr_h.csv};
\addplot[color=black, mark=+] table[
x=hidden_dim, y=memory, col sep=comma,
] {data/spectr_h.csv};

\nextgroupplot[xlabel=Number of Adapters, ylabel=FLOPs, xmode=log,ymode=log]
\addplot[color=orange, mark=*] table[
x=num_adapters, y=flops, col sep=comma,
] {data/arrow_n.csv};
\addlegendentry{\arrow}
\addplot[color=teal, mark=triangle*] table[
x=num_adapters, y=flops, col sep=comma,
] {data/lag_n.csv};
\addlegendentry{\lag}
\addplot[color=blue, mark=square*] table[
x=num_adapters, y=flops, col sep=comma,
] {data/seqr_n.csv};
\addlegendentry{\seqr}
\addplot[color=black, mark=+] table[
x=num_adapters, y=flops, col sep=comma,
] {data/spectr_n.csv};
\addlegendentry{\spectr}

\nextgroupplot[xlabel=Number of Adapters, ylabel=Memory, xmode=log,ymode=log]
\addplot[color=orange, mark=*] table[
x=num_adapters, y=memory, col sep=comma,
] {data/arrow_n.csv};
\addplot[color=teal, mark=triangle*] table[
x=num_adapters, y=memory, col sep=comma,
] {data/lag_n.csv};
\addplot[color=blue, mark=square*] table[
x=num_adapters, y=memory, col sep=comma,
] {data/seqr_n.csv};
\addplot[color=black, mark=+] table[
x=num_adapters, y=memory, col sep=comma,
] {data/spectr_n.csv};
\nextgroupplot[xlabel=LoRA Rank, ylabel=FLOPs,ymode=log, xmode=log, xtick={8, 16, 32, 64, 128, 256}, log x ticks with fixed point]
\addplot[color=orange, mark=*] table[
x=rank, y=flops, col sep=comma,
] {data/arrow_r.csv};
\addplot[color=teal, mark=triangle*] table[
x=rank, y=flops, col sep=comma,
] {data/lag_r.csv};
\addplot[color=blue, mark=square*] table[
x=rank, y=flops, col sep=comma,
] {data/seqr_r.csv};
\addplot[color=black, mark=+] table[
x=rank, y=flops, col sep=comma,
] {data/spectr_r.csv};

\nextgroupplot[xlabel=LoRA Rank, ylabel=Memory, xmode=log,ymode=log, xtick={8, 16, 32, 64, 128, 256}, log x ticks with fixed point]
\addplot[color=orange, mark=*] table[
x=rank, y=memory, col sep=comma,
] {data/arrow_r.csv};
\addplot[color=teal, mark=triangle*] table[
x=rank, y=memory, col sep=comma,
] {data/lag_r.csv};
\addplot[color=blue, mark=square*] table[
x=rank, y=memory, col sep=comma,
] {data/seqr_r.csv};
\addplot[color=black, mark=+] table[
x=rank, y=memory, col sep=comma,
] {data/spectr_r.csv};
\end{groupplot}
\end{tikzpicture}
    \caption{FLOPs (left) and GPU bytes used (right) for each method while varying hidden dimension (top), number of adapters in library (middle), and LoRA rank (bottom). Settings are fixed to $n=4096$, $N=1,000$, and $r=8$ when not under evaluation. \lag{} uses $k=20$ for \arrow{} filtering.}
    \label{fig:efficiency}
\end{figure}
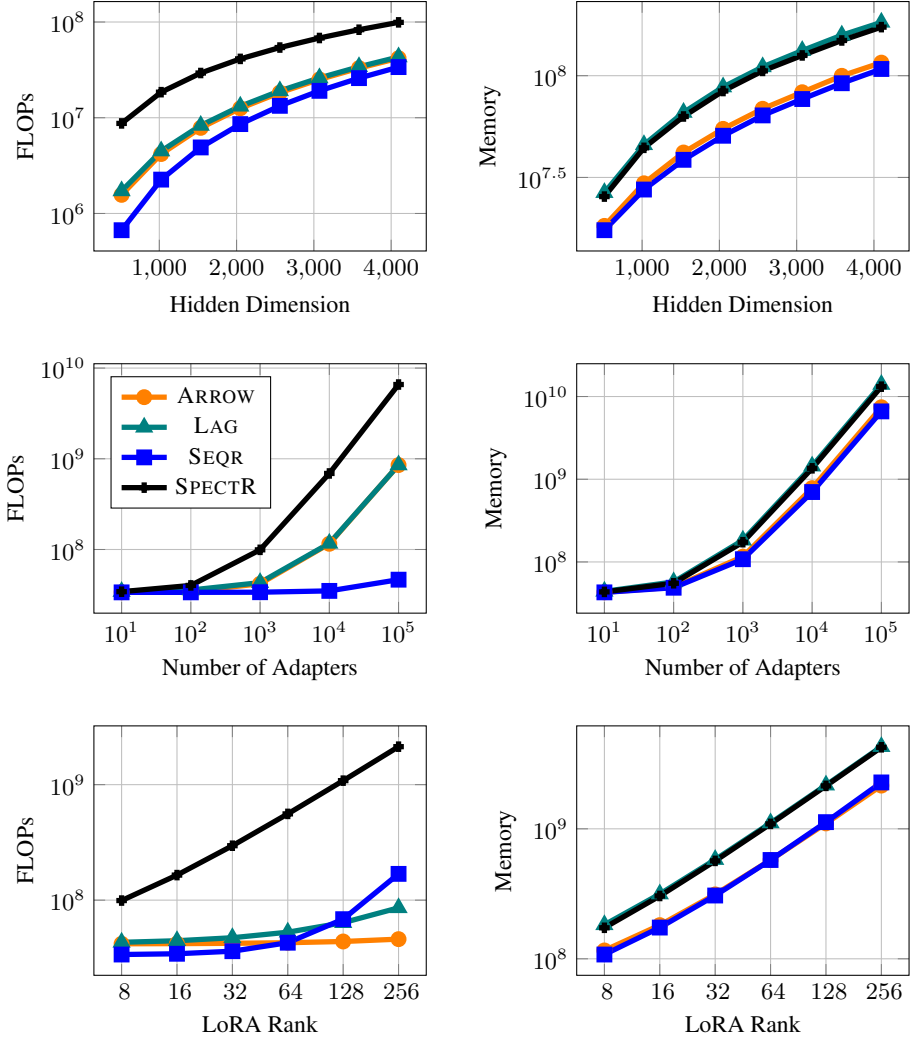

\seqr{} yields the same improved multi-task performance as \spectr{}, but with far greater efficiency. We measure the realized FLOPs and peak GPU memory used by each approach under various conditions (\autoref{fig:efficiency}). Total memory usage is dominated by the storage of the adapter library, so \seqr{} and \arrow{} are around twice as efficient when using shared $A$ matrices. \spectr{} and \lag{} require storing unique $\hat{A}_i$ matrices per adapter, even when the original $A$ matrix is shared. \seqr{} stores an extra $Nr^2$ parameters for the $R_i$ matrices while \arrow{} stores an extra $Nn$ for the arrow vectors. This gives \seqr{} an additional advantage in storage costs when $r^2 < n$. For computation, \seqr{} provides a significant reduction in FLOPs over other methods, especially for large adapter libraries using a smaller LoRA rank per adapter. \arrow{} requires fewer FLOPs than \seqr{} when $r > \sqrt{n}$, but the relative task-performance of \arrow{} degrades at higher rank, where routing decisions are still limited by the rank-1 prototypes \citep{spectr}.

\section{Conclusion}

In conclusion, we introduced \seqr{}, a state-of-the-art unsupervised LoRA routing algorithm. We formalized the goal of unsupervised LoRA routing in terms of activation norm-maximization and provided theoretical results for previous routing methods under this framework. The approaches that guarantee selecting the norm-maximizing adapter had better multi-task performance in our experiments. We showed that \seqr{} has this guarantee while being orders of magnitude more efficient than existing alternatives. \seqr{} leverages prior work showing that similar performance can be achieved when using LoRAs with frozen $A$ matrices shared across adapters, a finding we empirically validate. Sharing the $A$ matrices resulted in a higher variance in activation norms, which we corrected via an offline calibration step. Calibration improved performance for \spectr{} and \seqr{}, with \seqr{} being significantly faster in execution due to the increased efficiency. \seqr{} maintains the security benefits of other unsupervised methods, preventing data leakage without access to the LoRA weights.

\newpage



\bibliography{iclr2026_conference, anthology}
\bibliographystyle{iclr2026_conference}

\appendix

\section{Arrow Proof}
\label{construction}

\begin{proof}
    We will construct 2x2 LoRA adapters $C$ and $D$ and an input $\x$ that satisfy the condition of the theorem. 

    \begin{enumerate}
        \item Define $C=B_1A_1$:
        
        Let matrix $C = \begin{pmatrix} 2 & 0 \\ 0 & 1\end{pmatrix}$.

        The singular values are $\sigma_1(C) = 2$ and $\sigma_2(C) = 1$. The right singular vector corresponding to $\sigma_1$ is $\mathbf{v}_C = \begin{pmatrix}
            1 \\0
        \end{pmatrix}$.

        \item Define $D=B_2A_2$:

        We construct $D$ from the singular value decomposition $D=USV^T$ and choose the components to satisfy the theorem.

        Let $U=I$ the identity.

        Let the singular values be $\sigma_1(D)=3$ and $\sigma_2(D) = 1$.

        Let the right singular vectors be $\mathbf{v}_D = \frac{1}{\sqrt{2}}\begin{pmatrix}
            1\\1
        \end{pmatrix}$ and $\frac{1}{\sqrt{2}}\begin{pmatrix}
            1\\-1
        \end{pmatrix}$.

        $D = USV^T =  \begin{pmatrix}
            1 & 0 \\ 0 & 1
        \end{pmatrix}\begin{pmatrix}
            3 & 0 \\ 0 & 1
        \end{pmatrix}\begin{pmatrix}
            \frac{1}{\sqrt{2}} & \frac{1}{\sqrt{2}} \\ \frac{1}{\sqrt{2}} & -\frac{1}{\sqrt{2}}
        \end{pmatrix}^T = \begin{pmatrix}
            \frac{3}{\sqrt{2}} & \frac{3}{\sqrt{2}} \\ \frac{1}{\sqrt{2}} & -\frac{1}{\sqrt{2}}
        \end{pmatrix}$.

        \item Choose vector $\x$:

        Let $\x$ = $\mathbf{v}_C = \begin{pmatrix}
            1\\0
        \end{pmatrix}$.

        \item Verify inequality:

        \begin{align*}
            LHS & = \mathrm{argmax}_i\lvert\mathbf{v}_i^T\x\rvert \\
            &= \mathrm{argmax}_i\{\lvert\mathbf{v}_C^T\x\rvert, \lvert\mathbf{v}_D^T\x\rvert\} \\
            &= \mathrm{argmax}_i\{1, \frac{1}{\sqrt{2}}\} = \text{(adapter 1)}.
        \end{align*}
        \begin{align*}
            RHS &= \mathrm{argmax}_i\lvert\lvert B_iA_i\x\rvert\rvert_2 \\
            &= \mathrm{argmax}_i\{\lvert\lvert C\x\rvert\rvert_2, \lvert\lvert D\x\rvert\rvert_2\} \\
            &= \mathrm{argmax}_i\{ \begin{Vmatrix}2\\0\end{Vmatrix}_2, \begin{Vmatrix}\frac{3}{\sqrt{2}}\\\frac{1}{\sqrt{2}}\end{Vmatrix}_2\} \\
            &=\mathrm{argmax}_i\{2, \sqrt{5}\} = \text{(adapter 2)}.
        \end{align*}

        $LHS \ne RHS$.
    \end{enumerate}
\end{proof}

\section{SpectR Proof}
\label{pr:spectr}
\begin{proof}
    Let $B\in \mathbb{R}^{m \times r}$, $A \in \mathbb{R}^{r \times n}$, and $\x \in \mathbb{R}^n$. So,
    \begin{align*}
        \lvert\lvert BA\x\rvert\rvert_2 &= \lvert\lvert USV^T\x\rvert\rvert_2 && \text{(Equation 1)} \\
        &=\lvert\lvert U\hat{A}\x\rvert\rvert_2 && \text{(Equation 3)}\\
        &=\sqrt{\lvert\lvert U\hat{A}\x\rvert\rvert_2^2} \\
        &=\sqrt{\x^T\hat{A}^TU^TU\hat{A}\x} && \text{(Definition of squared 2-norm)}\\
        &=\sqrt{\x^T\hat{A}^T\hat{A}\x} && \text{(Orthonormal columns} \implies U^TU=I)\\
        &=\sqrt{\lvert\lvert\hat{A}\x\rvert\rvert_2^2} \\
        &=\lvert\lvert\hat{A}\x\rvert\rvert_2 \\
    \end{align*}
\end{proof}

\section{SEQR Proof}
\label{pr:seqr}
\begin{proof}
    Let $B\in \mathbb{R}^{m \times r}$, $A \in \mathbb{R}^{r \times n}$, $\x \in \mathbb{R}^n$, and $B=QR$ from Equation 4. So,
    \begin{align*}
        \lvert\lvert BA\x\rvert\rvert_2 &= \lvert\lvert QRA\x\rvert\rvert_2 && \text{(substitution)} \\
        &=\sqrt{\lvert\lvert QRA\x\rvert\rvert_2^2} \\
        &=\sqrt{\x^TA^TR^TQ^TQRA\x} \\
        &=\sqrt{\x^TA^TR^TRA\x} && \text{(Orthonormal columns} \implies Q^TQ=I)\\
        &=\sqrt{\lvert\lvert RA\x\rvert\rvert_2^2} \\
        &=\lvert\lvert RA\x\rvert\rvert_2 \\
    \end{align*}
\end{proof}

\section{Adapter Details}
\label{adapters}

We fit LoRA adapters targeting all attention layers in the network (query, key, value, and output projection layers). We choose initial settings for the LoRAs using the \texttt{unsloth} hyperparameter guide.\footnote{\url{https://docs.unsloth.ai/get-started/fine-tuning-llms-guide/lora-hyperparameters-guide}} We use rank-32 adapters with a LoRA $\alpha = 64$ and dropout of 0.05. We train for two epochs using a cosine schedule with warm-up ratio of 5\% and a batch size of 8. We sweep learning rates in the set \{5e-6, 1e-5, 2e-5, 5e-5, 1e-4, 2e-4, 5e-4, 1e-3, 2e-3, 5e-3\} for each dataset, but share learning rates across random seeds. 

\end{document}